\documentclass{article}
\usepackage{ijcai21}

\usepackage{times}
\usepackage{soul}
\usepackage{url}
\usepackage[hidelinks]{hyperref}
\usepackage[utf8]{inputenc}
\usepackage[small]{caption}
\usepackage{graphicx}
\usepackage{amsmath}
\usepackage{amsthm}
\usepackage{booktabs}
\usepackage{algorithm}
\usepackage{algorithmic}
\usepackage{booktabs}

\usepackage{natbib}
\usepackage{todonotes}
\usepackage{multirow}
\usepackage{enumitem}
\usepackage{placeins}
\usepackage{amsmath}
\usepackage[arrow]{xy}
\usepackage[caption=false]{subfig}
\usepackage{tabularx}

\newlength\myindent
\newcommand{\dlcomment}[1]{}

\usepackage{comment}

\usepackage{tikz}

\title{
Applying the Case Difference Heuristic \\ to Learn Adaptations from Deep Network Features}

\author{
Xiaomeng Ye \and
Ziwei Zhao\and
David Leake\and
Xizi Wang\And
David Crandall\\
\affiliations
Luddy School of Informatics, Computing, and Engineering \\       
  Indiana University, Bloomington IN 47408, USA \\             
\emails
\{xiaye, zz47, leake, xiziwang, djcran\}@indiana.edu
}

\begin{document}

\maketitle

\begin{abstract}
The case difference heuristic (CDH) approach is a knowledge-light method for learning case adaptation knowledge from the case base of a case-based reasoning system. Given a pair of cases, the CDH approach attributes the difference in their solutions to the difference in the problems they solve, and generates adaptation rules to adjust solutions accordingly when a retrieved case and new query have similar problem differences. As an alternative to learning adaptation rules, several researchers have applied neural networks to learn to predict solution differences from problem differences.
Previous work on such approaches has assumed that the feature set describing problems is predefined.  This paper investigates a two-phase process combining deep learning for feature extraction and neural network based adaptation learning from extracted features.  
Its performance is demonstrated in a regression task on an image data:  predicting age given the image of a face.  Results show that the combined process can successfully learn adaptation knowledge applicable to nonsymbolic differences in cases. The CBR system achieves slightly lower performance overall than a baseline deep network regressor, but better performance than the baseline on novel queries.
\end{abstract}

\section{Introduction}

A case-based reasoning (CBR) system solves new queries by retrieving a similar case from the case base (e.g., \cite{aamodt-plaza94,kolodner93,leake96-cbr-overview,mantaras-et-al05,riesbeck-schank89,richter-weber13}).  If the solution of the retrieved case  does not apply to the query, then an adaptation process modifies the solution to respond to situation differences. After the query is successfully solved, it is stored as a new case in the case base for future use. While a case base with good coverage and a good retrieval model allows the CBR system to retrieve similar cases, the case adaptation model determines the flexibility of the system to adjust the retrieved solutions for novel queries.

From the early days of CBR, adaptation has often been done using expert knowledge encoded in hand-crafted adaptation rules (e.g., \cite{hammond89}). 
To alleviate the burden of  knowledge engineering, the case difference heuristic (CDH) approach extracts case adaptation knowledge from cases in the case base \citep{hanney-keane96}.
The CDH approach collects pairs of cases from the case base, and generates rules that attribute the difference in the problem descriptions (the problem difference) of a pair to the difference in their solution descriptions (the solution difference).  The problem difference determines the antecedent of the new adaptation rule, and the solution difference determines its consequent. The resulting rules have the following form: If the problem difference between a query Q and the problem of a retrieved case C matches the problem difference of a rule R, then the solution difference of R can be applied to the solution of the retrieved case C.  For regression tasks, the consequent might be a numeric change to be applied to the predicted value of a retrieved case by addition, multiplication, or more complicated means. Furthermore, multiple rules can be generalized into one if they share similar preconditions and effects \citep{hanney-keane96}.

Deep learning (DL), using deep neural  networks and often learning from massive amounts of data, has shown the ability to extract useful features from nonsymbolic data.  It has been highly successful in many task domains, especially those resistant to other AI methods, such as computer vision tasks \citep{sinha2018deep}.

The CBR community has brought neural networks to CDH, using neural networks to learn adaptation knowledge from pairs of cases \citep{Policastro2003,POLICASTRO200626,liao-liu-chao18,craw-wiratunga-rowe06,leakeYe2021}. Up to this point, such approaches have used network learning to learn adaptations based on differences in sets of predefined features.   They have not attempted to exploit one of the noteworthy strengths of deep learning:  the ability to learn useful features from data.

This study extends previous work on a neural network based CDH method, NN-CDH \citep{leakeYe2021} by using a deep neural network to extract features, which are used both for retrieval and as inputs to adaptation knowledge learning using NN-CDH. From the perspective of CBR, the novel contribution of this paper is to use DL-generated features for adaptation learning.  From the perspective of DL, the novel contribution is to use CBR-style adaptation to handle novel queries.
The proposed method, deep neural network based CDH (DL-CDH), is used in a CBR system for a computer vision task of predicting the age of a person from their facial image. System performance is compared to that of a standard DL regression model and a retrieval-only system.

By using DL-CDH, our CBR system improves the solution provided by its retrieval stage using adaptation knowledge that does not require engineered features. It achieves slightly lower accuracy than a counterpart DL regressor, while carrying benefits of CBR such as explanation by cases and lazy incremental learning. Moreover, the CBR system can outperform the DL regressor for out-of-distribution queries by using adaptation knowledge learned by DL-CDH.

\section{Background}

\subsection{Image Processing using CBR}\label{sec:ImageInCBR}

CBR researchers have been tackling computer vision problems for some time. \cite{perner01} describes image interpretation as a multi-level process, from pre-processing pixels at low levels, to segmenting and extracting features at intermediate levels, to classification and understanding at high levels. She proposes using CBR on different levels, mostly focusing on the case representation of images and the similarity measure between cases.
A series of works follows this line of research \citep{perner-holt-richter05,wilson-osullivan08,perner2017}. 
In a recent survey \cite{perner2017} reviews applications of CBR in parameter selection, image interpretation, incremental prototype-based classification, novelty detection, and 1-D signal representation.

A more recent survey of CBR work \citep{Rahul2020} reports on multiple works on CBR image processing systems following a framework in which features extracted from images and their class labels are combined into cases upon which CBR systems operate. 
This framework can be considered as a simpler version of that in \cite{perner01}. The research projects surveyed in this study focus on feature extraction and case retrieval for a wide range of domains.

Object detection and classification is a more direct problem in image interpretation. \cite{Turner2018} use a Convolutional Neural Network (CNN) to extract features and identify parts. Features and parts are used as a case's problem description and the object label is the solution description. In contrast to the work in \cite{Rahul2020}, they perform novel object discovery using CBR,  by retaining a case if the system deems the case novel, and retrieving and reusing this new case later. \cite{Turner2019} further extend this approach by integrating it with a CNN, so that familiar queries are handled by a CNN while novel queries, when the CNN displays high uncertainty in its output, are handled by a CBR system.
The approach in this study is consistent with the framework by \cite{Rahul2020} and also uses a CNN for feature extraction as \cite{Turner2018}, adding learned case adaptation.

\subsection{Siamese Networks as a Similarity Measure}

A Siamese network takes two input vectors, extracts features from each input using the same feature extraction network, passes the extracted features into a distance layer, and calculates the distance between them \citep{bromley93,Bromley:1993:SVU:2987189.2987282}.  Similar to \citet{Martin2017ACS} and \citet{mathisen-et-al19}, this study uses a Siamese network as a similarity measure for case retrieval in CBR.

 \cite{wetzel2020twin} use a Siamese network to predict the target value differences given two data points. This approach can predict a target value of a query by pairing it with a data point, predicting the target value difference, and adding it back to the target value of the data point. Their method predicts a target value by projecting from an ensemble of training data points. This is in spirit very similar to the case difference heuristic methods described in the following section.
 
\subsection{Adaptation Based on Features Extracted by Neural Networks}

For the purpose of generating semi-factual and counterfactual case explanations, \cite{kenny2020} use a convolutional neural network to extract features, identify and modify exceptional features, and visualize the modified feature as an image using a generative adversarial network (GAN). Their work adapts the features generated by DL, while our work is adapting case solutions.  

\subsection{Neural Network Based Applications of the Case Difference Heuristic}

Much research has investigated symbolic AI methods for reasoning about case differences to  generate case adaptation rules ({\it e.g.}, \cite{hanney-keane96,jalali-Leake13-2,McDonnell-Cunningham06,McSherry98,wilke-etal97,aquin-et-al07,craw-wiratunga-rowe06}). 
 Of particular interest here is how network methods have been used for case difference heuristic processes. 
\cite{Policastro2003,POLICASTRO200626} and \cite{liao-liu-chao18} use neural networks to learn relationships between problem differences and solution differences, and use these networks to predict a solution difference from a problem difference.

Inspired by previous work in network-based CDH,
\cite{leakeYe2021} develops a neural network based CDH approach, NN-CDH, in which a neural network learns to predict solution differences for regression problems based on the context of adaptation and the problem difference. NN-CDH has shown good results in improving retrieval results.  In addition, in initial tests for scenarios in which when domain knowledge is hard to learn, such as novel queries in high dimensionality, the CBR system using NN-CDH for adaptation could outperform a baseline neural network regressor.  Extending the work in \cite{leakeYe2021}, this study:
\begin{enumerate}
    \item Uses features extracted by a deep neural network as problem descriptions of cases and learns adaptation knowledge from extracted features;
    \item Evaluates the effects of CDH adaptations on a CBR system with two different retrieval methods.
    \item Compares the performance of the CBR system with its counterpart DL system (instead of a neural network system which is compared with NN-CDH) in an image domain task (instead of tabular data domains).
\end{enumerate}

\subsection{Age Prediction from Facial Images using Deep Learning}

This study examines the age prediction task involving out-of-distribution (OOD) samples. Age prediction is a well-studied topic in computer vision. Interestingly it is tackled both as a classification problem \citep{LH:CVPRw15:age,DUAN2018448} and as a regression problem \citep{7406403}. An ensemble attentional convolutional network is proposed in \cite{abdolrashidi2020age} to accomplish both age and gender prediction tasks. \cite{9144212} survey the detection of OOD data in DL, but handling of OOD data is one step beyond detection. To handle biases in age, ethnicity, and gender prediction, \cite{cao2021outofdistribution} proposed distribution-aware techniques in data augmentation and data curation to ensure fairness.

This study does not try to surpass state-of-art age prediction studies (see \cite{cao2021outofdistribution}). Instead we chose this task to test the effectiveness of the DL-CDH approach and the benefit of CDH adaptation in DL tasks involving OOD samples.


\section{A Deep Learning Based Case Difference Heuristic Approach}

Similarly to its predecessor NN-CDH, DL-CDH learns adaptation knowledge by training a neural network over pairs of cases to predict solution difference based on problem difference (and adaptation context). In our testbed system, DL-CDH is implemented as a feedforward neural network with dense layers and dropout layers. However, the adaptation network of DL-CDH does not learn directly from the problem descriptions of the cases.  Instead, the adaptation network learns using features extracted from a deep neural network for problem  representations. Given a pair of cases, the network takes the pair of extracted features as input and outputs the predicted age difference between the two cases.


A CBR system solving DL regression tasks can use DL-CDH as its adaptation process. Given a query, the CBR system first retrieves a case. Then a deep neural network extracts the features of the query and the retrieved case. DL-CDH uses their features to predict the solution difference between the query and the retrieved case. Last, this solution difference is applied to the solution of the retrieved case to yield the final solution.

\section{Evaluation}

\subsection{Dataset and Preprocessing}

\begin{figure}[]
    \centering
    \includegraphics[width=0.45\textwidth]{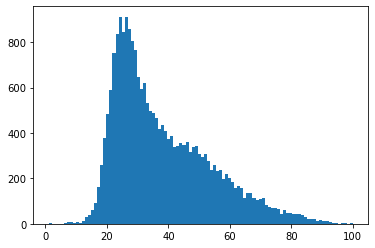}
    \caption{Age distribution of wiki dataset}
    \label{fig:agedistribution}
\end{figure}

We evaluate the perfomance of DL-CDH for the task of predicting the age of a person given the image of a face. 
We used Wikipedia images from the IMDB-WIKI dataset \citep{Rothe-IJCV-2018}. We filtered out all images with no face or more than one face, yielding a new dataset containing 22578 face images, each of which is a $224\times224$ image with RGB values per pixel, labeled with an age ranging from 1 to 100. The age distribution is shown in Figure \ref{fig:agedistribution}.

We used a pretrained CNN with the vgg-vd-16 \citep{simonyan2015deep} architecture to extract features from face images. The CNN is pretrained on the vgg-face dataset \citep{Parkhi15} for face recognition task.
We pass all images to the CNN and convert each image $x$ into a case with a feature vector of $2622$ dimensions and associated with a solution label, a numeric value $sol(x)$ representing the age of the image subject. We refer to the feature extractor function (VGG-Face CNN) as $f$, and the extracted features of a case $x$ is $f(x)= (x_1, x_2 ..., x_{2622})$.

The data set is prepared for experiments in two settings:
\begin{enumerate}
    \item Normal Setting: The cases are used in a 10-fold cross-validation, where 80\% of the cases are used for training, 10\% for validation and 10\% for testing.
    \item Novel Query Setting: The cases of age 20-50 are used in a 10-fold cross-validation with 90\% cases for training and 10\% for validation, while cases of age 0-20, 50-70, and $>$70 are respectively used as testing queries in multiple experiments.
\end{enumerate}

\subsection{Experimental Settings}

The testbed system involves only a retrieval stage and an adaptation stage. Because the goal is to study the effectiveness of adaptation, retention effects are not considered.

\begin{figure*}[tb]
    \centering
    \includegraphics[width=\textwidth]{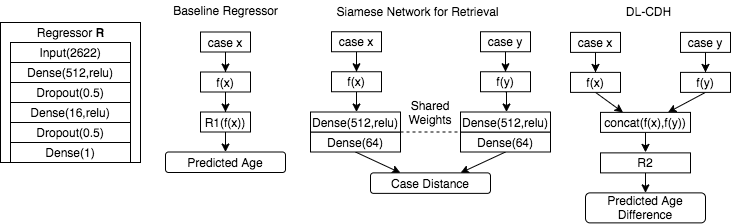}
    \caption{Network Architectures of components of the Testbed System. Left: Base Regression Network Used in Regressor and DL-CDH. Center Left: Baseline Regressor. Center Right: Siamese Network Used to Retrieve Cases of Similar Ages. Right: DL-CDH Used to Adapt Initial Age Prediction from Retrieval}
    \label{fig:network}
\end{figure*}

\subsubsection{Retrieval Method}

We implemented two similarity measures for retrieval. The first is a 1-nearest neighbor search over the extracted features $f(x)= (x_1, x_2 ...)$, where all features are weighted equally. The distance between two cases is calculated as the sum of feature-wise L1 distances.

The second similarity measure is a Siamese network trained on triplets of cases from the case base. 
The network's architecture is depicted in center right of Figure \ref{fig:network}. The network is trained by triplets of $(a, p, n)$ generated on the fly: given an anchor case $a$, a positive case $p$ is another case chosen from cases of the same age as anchor $a$, and a negative case $n$ is another case chosen from cases whose age is at least 10 years different from that of anchor $a$.
The network is trained using the triplet margin loss:
\begin{equation*}
    L(a,p,n) = \max(d(a_i, p_i)- d(a_i, n_i)+ margin, 0)
\end{equation*}
where $d(x_i,y_i) = ||x_i - y_i||_p$. In our implementation, we set $margin=1$ and $p=1$.


\subsubsection{Adaptation Method}

The network architecture of the testbed DL-CDH $R2$ is depicted in the right of Figure \ref{fig:network}. We trained DL-CDH with pairs of cases generated on the fly: For every training case $x$, another random training case $y$ is paired. 
We concatenate $f(x)$ and $f(y)$ into a single vector $concat(f(x),f(y))$ as input to represent their difference, and the age difference becomes the expected output $sol(x)-sol(y)$.  For the purpose of validation, every validation case in the validation set is paired with its nearest neighbor (under L1 distance) in the training case, and all such pairs form the validation pair set for the training of DL-CDH. 

\subsubsection{Baseline Models}

For comparison, we implemented a neural network regressor $R1$ that directly predicts solution $sol(x)$ based on case features $f(x)$. The architecture is depicted in the center left of Figure \ref{fig:network}.  To ensure fairness of the comparison, the baseline regressor $R1$ uses the same inner architecture as $R2$ in DL-CDH (as shown in the left of Figure \ref{fig:network}). In other words, $R1$ and $R2$ share the same layers, number of neurons, and activation functions, but they are trained for different purposes: $R1$ learns to predict age $sol(x)$ from feature set $f(x)$, while $R2$ learns to predict age difference $sol(x)-sol(y)$ from feature difference $concat(f(x),f(y))$. The regressor is trained until its error on the validation case set converges.
Last, a constant baseline is implemented by making predictions using the average label of all training samples.

\vspace{\baselineskip}

In summary, four systems are compared in the experiments: a constant baseline, baseline regressor, CBR system with L1 distance as similarity measure and DL-CDH as adaptation (referred as ``L1 + adaptation''), and CBR system with Siamese network as similarity measure and DL-CDH as adaptation (referred as ``Siamese + adaptation''). The experimental source code is available online\footnote{https://github.com/ziweizhao1993/DL-CBH}.
All systems are trained for 50 epochs using the Adam optimizer \citep{kingma2017adam} and a learning rate of $10^{-4}$. After the training of each system, the model with the highest validation accuracy is selected for testing. 

\subsection{Experimental Results}

The testing errors of systems under different settings are shown in Table \ref{tab:results}. For the CBR systems, the error of initial solution by retrieval  and the error of final solution after adaptation are both shown.

\begin{table*}[t]
\centering
\begin{tabular}{lcccccccc}
\multirow{2}{*}{} & \multirow{2}{*}{\begin{tabular}[c]{@{}c@{}}Training\\ Age Range\end{tabular}} & \multirow{2}{*}{\begin{tabular}[c]{@{}c@{}}Query\\ Age Range\end{tabular}} & \multicolumn{2}{c}{Baseline} & \multicolumn{2}{c}{L1 + Adaptation} & \multicolumn{2}{c}{Siamese + Adaptation} \\ \cmidrule(lr){4-5} \cmidrule(lr){6-7} \cmidrule(lr){8-9}
 &  &  & Constant & Regressor & Retrieve & Adapt & Retrieve & Adapt \\ \midrule 
\multicolumn{1}{l}{\begin{tabular}[c]{@{}l@{}}Normal\\ Setting\end{tabular}} & All & All & 12.94 & \textbf{5.9668} & 9.5805 & 8.1397 & 7.4944 & 7.8558 \\ \midrule 
\multicolumn{1}{l}{\multirow{3}{*}{\begin{tabular}[c]{@{}l@{}}Novel \\ Query \\ Setting\end{tabular}}} & \multirow{3}{*}{20$\sim$50} & \textless{}20 & 14.7184 & 8.948 & 10.1784 & 9.1764 & 8.6156 & \textbf{8.0906} \\
\multicolumn{1}{l}{} &  & 50$\sim$70 & 25.8215 & 15.5101 & 17.3237 & 14.7365 & 14.8493 & \textbf{12.0175} \\
\multicolumn{1}{l}{} &  & \textgreater{}70 & 45.8084 & 32.7671 & 35.8406 & 32.1052 & 33.049 & \textbf{28.8334} \\
\midrule 
\end{tabular}
\caption{Average Error of Systems under Different Settings}
\label{tab:results}
\end{table*}

\subsubsection{Effects of the Adaptation by DL-CDH over Retrieval}

By comparing the errors before and after adaptation in the CBR systems, we observe that adaptation in general improves the initial solution by retrieval, with only one exception: In ``Siamese + adaptation'' under normal settings, on average the adaptation actually provides worse solutions compared to the initial solution by retrieval. 

This is a phenomenon also observed and explained in \cite{LeakeYe2021underR}. Because retrieval and adaptation are trained independently, the two stages may be out of synchronization. In the case of ``Siamese + adaptation'' under normal setting, the retrieval stage is well trained to retrieve cases that are close to queries, while the adaptation stage is trained on random pairs and capable of adapting for relatively larger difference between case pairs. The adaptation stage is not trained to handle the pairs of query and retrieved cases and is therefore not guaranteed to improve the initial solution of retrieval. Adaptation-guided retrieval might be a way to alleviate this problem. \cite{LeakeYe2021underR} further studies this phenomenon and proposes an alternating optimization process in response.

In ``L1 + adaptation'' or in ``Siamese + adaptation'' under novel query settings, the retrieved cases are not as close to the queries and the adaptation stage is more suited to handle such adaptation. Therefore the initial solutions are consistently improved by adaptation.

\subsubsection{Comparison between Baseline Regressor and CBR System}

In the normal setting, the baseline regressor is the best performing system. The CBR systems perform worse than the regressor but better than the constant baseline. With abundant data, the regressor consistently outperforms its counterpart CBR systems because the regressor network is capable of learning the domain knowledge well and handling non-novel queries.  This is not surprising given the quality of performance achieved by deep learning provided with sufficient data.

In the novel query setting, ``L1 + adaptation'' performs similarly to or even better than the baseline regressor. Additionally, ``Siamese + adaptation" further improves upon ``L1 + adaptation''  and is the best performing of all systems. As novel queries become harder to solve (further from the training case distribution), the benefit of ``Siamese + adaptation" increases. This shows that adaptation knowledge learned by DL-CDH indeed adapts the initial solution by retrieval to better solve the queries, and if the initial solution is closer to the real solution, the adaptation also tends to be more accurate.

The comparison between the baseline regressor and the CBR system is consistent with the results and discussion in \cite{leakeYe2021}, where NN-CDH, the predecessor of DL-CDH, performs worse than its counterpart regressor in an easy task domain but better than the regressor in a domain with  novel queries. It also matches with the motivation of \cite{Turner2019}, where familiar queries are handled by a CNN, but novel queries are handled by a CBR system.

\section{Future Work}

\subsection{Extending to Classification Task Domains}

NN-CDH has been extended to learn adaptation knowledge in classification task domains \citep{ye-et-al21}.
As a descendent of NN-CDH, DL-CDH can also be extended for classification. As discussed in Section \ref{sec:ImageInCBR}, CBR has been applied in many image classification tasks using DL techniques, but the adaptation stage is not a focus in any existing research. We envision that DL-CDH, once extended for classification, can be a natural extension for works such as \cite{Turner2018,Turner2019}.

\subsection{Image Adaptation Using DL-CDH and GAN}

In work highly relevant to this study, \cite{kenny2020} propose modifying extracted DL features to generate  counterfactual and semi-factual cases. They take the additional step of recovering image from the modified features by using a generative adversarial network. Inspired by their work, we consider the features modified by DL-CDH as candidates to project back to image space using a GAN. 

In contrast to current models in adversarial-based adaptation \citep{DBLP:journals/corr/abs-1812-04948,DBLP:journals/corr/abs-1711-10678}, whose adaptations are trained and used in a network doing end-to-end processing, 
adaptations in DL-CDH are more localized; consequently, they can be more transparent and subject to manual control. 

\section{Conclusion}

This study extends the neural network based case difference heuristic approach by combining it with feature extraction from a deep neural network. Its performance was evaluated on a facial image data set for the task of age prediction.
Except for one  situation in which retrieval and adaptation are out-of-synchronization, adaptation by DL-CDH consistently improves the initial solutions provided by retrieval. The testbed DL-CDH system consistently outperforms its counterpart neural network regressor when solving novel queries, but it still loses to its counterpart for queries within the distribution of the extensive training data. However, we note that the CBR system offers benefits beyond accuracy alone, such as lazy learning and explanation by cases, enabling online learning without costly retraining.
The work suggests future directions in extending DL-CDH for classification and combining it with GANs to adapt images. 

\section*{Acknowledgement}

We acknowledge support from the Department of the Navy, Office of Naval Research (Award N00014-19-1-2655), and the
US Department of Defense (Contract W52P1J2093009).

\bibliographystyle{named}
\bibliography{main}

\begin{thebibliography}{}

\bibitem[\protect\citeauthoryear{Aamodt and Plaza}{1994}]{aamodt-plaza94}
A.~Aamodt and E.~Plaza.
\newblock Case-based reasoning: Foundational issues, methodological variations,
  and system approaches.
\newblock {\em \uppercase{AI} Communications}, 7(1):39--52, 1994.

\bibitem[\protect\citeauthoryear{Abdolrashidi \bgroup \em et al.\egroup
  }{2020}]{abdolrashidi2020age}
A.~Abdolrashidi, M.~Minaei, E.~Azimi, and S.~Minaee.
\newblock Age and gender prediction from face images using attentional
  convolutional network.
\newblock {\em arXiv preprint arXiv:2010.03791}, 2020.

\bibitem[\protect\citeauthoryear{Barman \bgroup \em et al.\egroup
  }{2020}]{Rahul2020}
R.~Barman, S.~Kr. Biswas, S.~Sarkar, B.~Purkayastha, and B.~Soni.
\newblock Image processing using case-based reasoning: A survey.
\newblock In Pradeep~Kumar Mallick, Preetisudha Meher, Alak Majumder, and
  Santos~Kumar Das, editors, {\em Electronic Systems and Intelligent
  Computing}, pages 653--661, Singapore, 2020. Springer Singapore.

\bibitem[\protect\citeauthoryear{Bromley \bgroup \em et al.\egroup
  }{1993a}]{bromley93}
J.~Bromley, J.~W. Bentz, L.~Bottou, I.~Guyon, Y.~LeCun, C.~Moore, E.~Sackinger,
  and R.~Shah.
\newblock Signature verification using a siamese time delay neural network.
\newblock {\em International Journal of Pattern Recognition and Artificial
  Intelligence}, 7(04):669--688, 1993.

\bibitem[\protect\citeauthoryear{Bromley \bgroup \em et al.\egroup
  }{1993b}]{Bromley:1993:SVU:2987189.2987282}
J.~Bromley, I.~Guyon, Y.~LeCun, E.~S\"{a}ckinger, and R.~Shah.
\newblock Signature verification using a "siamese" time delay neural network.
\newblock In {\em Proceedings of the 6th International Conference on Neural
  Information Processing Systems}, NIPS'93, pages 737--744, San Francisco, CA,
  USA, 1993. Morgan Kaufmann Publishers Inc.

\bibitem[\protect\citeauthoryear{Bulusu \bgroup \em et al.\egroup
  }{2020}]{9144212}
S.~Bulusu, B.~Kailkhura, B.~Li, P.~K. Varshney, and D.~Song.
\newblock Anomalous example detection in deep learning: A survey.
\newblock {\em IEEE Access}, 8:132330--132347, 2020.

\bibitem[\protect\citeauthoryear{Cao \bgroup \em et al.\egroup
  }{2021}]{cao2021outofdistribution}
Y.~Cao, D.~Berend, P.~Tolmach, G.~Amit, M.~Levy, Y.~Liu, A.~Shabtai, and
  Y.~Elovici.
\newblock Out-of-distribution detection and generalization to enhance fairness
  in age prediction, 2021.

\bibitem[\protect\citeauthoryear{Duan \bgroup \em et al.\egroup
  }{2018}]{DUAN2018448}
M.~Duan, K.~Li, C.~Yang, and K.~Li.
\newblock A hybrid deep learning cnn–elm for age and gender classification.
\newblock {\em Neurocomputing}, 275:448--461, 2018.

\bibitem[\protect\citeauthoryear{D'\uppercase{A}quin \bgroup \em et al.\egroup
  }{2007}]{aquin-et-al07}
M.~D'\uppercase{A}quin, F.~Badra, S.~Lafrogne, J.~Lieber, A.~Napoli, and
  L.~Szathmary.
\newblock Case base mining for adaptation knowledge acquisition.
\newblock In {\em Proceedings of the Twentieth International Joint Conference
  on Artificial Intelligence (IJCAI-07)}, pages 750--755, San Mateo, 2007.
  Morgan Kaufmann.

\bibitem[\protect\citeauthoryear{Hammond}{1989}]{hammond89}
K.~Hammond.
\newblock {\em Case-Based Planning: Viewing Planning as a Memory Task}.
\newblock Academic Press, San Diego, 1989.

\bibitem[\protect\citeauthoryear{Hanney and Keane}{1996}]{hanney-keane96}
K.~Hanney and M.~Keane.
\newblock Learning adaptation rules from a case-base.
\newblock In {\em Proceedings of the Third European Workshop on Case-Based
  Reasoning}, pages 179--192, Berlin, 1996. Springer.

\bibitem[\protect\citeauthoryear{He \bgroup \em et al.\egroup
  }{2017}]{DBLP:journals/corr/abs-1711-10678}
Z.~He, W.~Zuo, M.~Kan, S.~Shan, and X.~Chen.
\newblock Arbitrary facial attribute editing: Only change what you want.
\newblock {\em CoRR}, abs/1711.10678, 2017.

\bibitem[\protect\citeauthoryear{Jalali and Leake}{2013}]{jalali-Leake13-2}
V.~Jalali and D.~Leake.
\newblock Extending case adaptation with automatically-generated ensembles of
  adaptation rules.
\newblock In {\em Case-Based Reasoning Research and Development,
  \uppercase{ICCBR} 2013}, pages 188--202, Berlin, 2013. Springer.

\bibitem[\protect\citeauthoryear{Karras \bgroup \em et al.\egroup
  }{2018}]{DBLP:journals/corr/abs-1812-04948}
T.~Karras, S.~Laine, and T.~Aila.
\newblock A style-based generator architecture for generative adversarial
  networks.
\newblock {\em CoRR}, abs/1812.04948, 2018.

\bibitem[\protect\citeauthoryear{{Kenny} and {Keane}}{2020}]{kenny2020}
E.~M. {Kenny} and M.~T. {Keane}.
\newblock {On Generating Plausible Counterfactual and Semi-Factual Explanations
  for Deep Learning}, 2020.
\newblock arXiv:2009.06399.

\bibitem[\protect\citeauthoryear{Kingma and Ba}{2017}]{kingma2017adam}
D.~P. Kingma and J.~Ba.
\newblock Adam: A method for stochastic optimization, 2017.
\newblock arXiv:1412.6980.

\bibitem[\protect\citeauthoryear{Kolodner}{1993}]{kolodner93}
J.~Kolodner.
\newblock {\em Case-Based Reasoning}.
\newblock Morgan Kaufmann, San Mateo, CA, 1993.

\bibitem[\protect\citeauthoryear{Leake and Ye}{2021}]{LeakeYe2021underR}
D.~Leake and X.~Ye.
\newblock Harmonizing case retrieval and adaptation with alternating
  optimization, 2021.
\newblock In press.

\bibitem[\protect\citeauthoryear{Leake \bgroup \em et al.\egroup
  }{2021}]{leakeYe2021}
D.~Leake, X.~Ye, and D.~Crandall.
\newblock Supporting case-based reasoning with neural networks: An illustration
  for case adaptation.
\newblock 2021.

\bibitem[\protect\citeauthoryear{Leake}{1996}]{leake96-cbr-overview}
D.~Leake.
\newblock \uppercase{CBR} in context: The present and future.
\newblock In D.~Leake, editor, {\em Case-Based Reasoning: Experiences, Lessons,
  and Future Directions}, pages 3--30. \uppercase{AAAI} Press, Menlo Park,
  \uppercase{CA}, 1996.
\newblock http://www.cs.indiana.edu/\~{}leake/papers/a-96-01.html.

\bibitem[\protect\citeauthoryear{Levi and Hassner}{2015}]{LH:CVPRw15:age}
G.~Levi and T.~Hassner.
\newblock Age and gender classification using convolutional neural networks.
\newblock In {\em IEEE Conf. on Computer Vision and Pattern Recognition (CVPR)
  workshops}, June 2015.

\bibitem[\protect\citeauthoryear{Liao \bgroup \em et al.\egroup
  }{2018}]{liao-liu-chao18}
C.~Liao, A.~Liu, and Y.~Chao.
\newblock A machine learning approach to case adaptation.
\newblock {\em 2018 IEEE First International Conference on Artificial
  Intelligence and Knowledge Engineering (AIKE)}, pages 106--109, 2018.

\bibitem[\protect\citeauthoryear{{L\'{o}pez de M\'{a}ntaras} \bgroup \em et
  al.\egroup }{2005}]{mantaras-et-al05}
R.~{L\'{o}pez de M\'{a}ntaras}, D.~McSherry, D.~Bridge, D.~Leake, B.~Smyth,
  S.~Craw, B.~Faltings, M.~Maher, M.~Cox, K.~Forbus, M.~Keane, A.~Aamodt, and
  I.~Watson.
\newblock Retrieval, reuse, revision, and retention in \uppercase{CBR}.
\newblock {\em Knowledge Engineering Review}, 20(3), 2005.

\bibitem[\protect\citeauthoryear{Martin \bgroup \em et al.\egroup
  }{2017}]{Martin2017ACS}
K.~Martin, N.~Wiratunga, S.~Sani, S.~Massie, and J.~Clos.
\newblock A convolutional siamese network for developing similarity knowledge
  in the selfback dataset.
\newblock In {\em ICCBR (Workshops)}, 2017.

\bibitem[\protect\citeauthoryear{Mathisen \bgroup \em et al.\egroup
  }{2019}]{mathisen-et-al19}
B.~M. Mathisen, A.~Aamodt, K.~Bach, and H.~Langseth.
\newblock Learning similarity measures from data.
\newblock {\em Progress in Artificial Intelligence}, 10 2019.

\bibitem[\protect\citeauthoryear{McDonnell and
  Cunningham}{2006}]{McDonnell-Cunningham06}
N.~McDonnell and P.~Cunningham.
\newblock A knowledge-light approach to regression using case-based reasoning.
\newblock In {\em Proceedings of the 8th European conference on Case-Based
  Reasoning}, ECCBR'06, pages 91--105, Berlin, 2006. Springer.

\bibitem[\protect\citeauthoryear{McSherry}{1998}]{McSherry98}
D.~McSherry.
\newblock An adaptation heuristic for case-based estimation.
\newblock In {\em Proceedings of the Fourth European Workshop on Advances in
  Case-Based Reasoning}, EWCBR '98, pages 184--195, London, 1998. Springer.

\bibitem[\protect\citeauthoryear{Parkhi \bgroup \em et al.\egroup
  }{2015}]{Parkhi15}
O.~M. Parkhi, A.~Vedaldi, and A.~Zisserman.
\newblock Deep face recognition.
\newblock In {\em British Machine Vision Conference}, 2015.

\bibitem[\protect\citeauthoryear{Perner \bgroup \em et al.\egroup
  }{2005}]{perner-holt-richter05}
P.~Perner, A.~Holt, and Michael Richter.
\newblock Image processing in case-based reasoning.
\newblock {\em Knowledge Engineering Review}, 20(3):311--314, 2005.

\bibitem[\protect\citeauthoryear{Perner}{2001}]{perner01}
P.~Perner.
\newblock Why case-based reasoning is attractive for image interpretation.
\newblock In {\em Case-Based Reasoning Research and Development: Proceedings of
  the Fourth International Conference on Case-Based Reasoning, ICCBR-01}, pages
  27--43, Berlin, 2001. Springer-Verlag.

\bibitem[\protect\citeauthoryear{Perner}{2017}]{perner2017}
P.~Perner.
\newblock Model development and incremental learning based on case-based
  reasoning for signal and image analysis.
\newblock In R.~P. Barneva, V.~E. Brimkov, and J.~M.~R.S. Tavares, editors,
  {\em Computational Modeling of Objects Presented in Images. Fundamentals,
  Methods, and Applications}, pages 3--24, Cham, 2017. Springer.

\bibitem[\protect\citeauthoryear{Policastro \bgroup \em et al.\egroup
  }{2003}]{Policastro2003}
C.~A. Policastro, A.~C. P. L.~F. Carvalho, and A.~C.~B. Delbem.
\newblock Hybrid approaches for case retrieval and adaptation.
\newblock In Andreas G{\"u}nter, Rudolf Kruse, and Bernd Neumann, editors, {\em
  KI 2003: Advances in Artificial Intelligence}, pages 297--311, Berlin,
  Heidelberg, 2003. Springer Berlin Heidelberg.

\bibitem[\protect\citeauthoryear{Policastro \bgroup \em et al.\egroup
  }{2006}]{POLICASTRO200626}
C.~Policastro, A.~Carvalho, and A.~Delbem.
\newblock Automatic knowledge learning and case adaptation with a hybrid
  committee approach.
\newblock {\em Journal of Applied Logic}, 4(1):26--38, 2006.

\bibitem[\protect\citeauthoryear{Ranjan \bgroup \em et al.\egroup
  }{2015}]{7406403}
R.~Ranjan, S.~Zhou, J.~C. Chen, A.~Kumar, A.~Alavi, V.~M. Patel, and
  R.~Chellappa.
\newblock Unconstrained age estimation with deep convolutional neural networks.
\newblock In {\em 2015 IEEE International Conference on Computer Vision
  Workshop (ICCVW)}, pages 351--359, 2015.

\bibitem[\protect\citeauthoryear{Richter and Weber}{2013}]{richter-weber13}
M.~Richter and R.~Weber.
\newblock {\em Case-Based Reasoning: A Textbook}.
\newblock Springer, Berlin, 2013.

\bibitem[\protect\citeauthoryear{Riesbeck and Schank}{1989}]{riesbeck-schank89}
C.~Riesbeck and R.C. Schank.
\newblock {\em Inside Case-Based Reasoning}.
\newblock Lawrence Erlbaum, Hillsdale, NJ, 1989.

\bibitem[\protect\citeauthoryear{Rothe \bgroup \em et al.\egroup
  }{2018}]{Rothe-IJCV-2018}
R.~Rothe, R.~Timofte, and L.~V. Gool.
\newblock Deep expectation of real and apparent age from a single image without
  facial landmarks.
\newblock {\em International Journal of Computer Vision}, 126(2-4):144--157,
  2018.

\bibitem[\protect\citeauthoryear{Simonyan and
  Zisserman}{2015}]{simonyan2015deep}
K.~Simonyan and A.~Zisserman.
\newblock Very deep convolutional networks for large-scale image recognition,
  2015.

\bibitem[\protect\citeauthoryear{Sinha \bgroup \em et al.\egroup
  }{2018}]{sinha2018deep}
R.~K. Sinha, R.~Pandey, and R.~Pattnaik.
\newblock Deep learning for computer vision tasks: a review.
\newblock {\em arXiv preprint arXiv:1804.03928}, 2018.

\bibitem[\protect\citeauthoryear{Turner \bgroup \em et al.\egroup
  }{2018}]{Turner2018}
J.~T. Turner, M.~W. Floyd, K.~M. Gupta, and D.~W. Aha.
\newblock Novel object discovery using case-based reasoning and convolutional
  neural networks.
\newblock In Michael~T. Cox, Peter Funk, and Shahina Begum, editors, {\em
  Case-Based Reasoning Research and Development}, pages 399--414, Cham, 2018.
  Springer International Publishing.

\bibitem[\protect\citeauthoryear{Turner \bgroup \em et al.\egroup
  }{2019}]{Turner2019}
J.~T. Turner, M.~Floyd, K.~Gupta, and T.~Oates.
\newblock Nod-cc: A hybrid cbr-cnn architecture for novel object discovery.
\newblock 06 2019.

\bibitem[\protect\citeauthoryear{Wetzel \bgroup \em et al.\egroup
  }{2020}]{wetzel2020twin}
S.~J. Wetzel, K.~Ryczko, R.~G. Melko, and I.~Tamblyn.
\newblock Twin neural network regression, 2020.

\bibitem[\protect\citeauthoryear{Wilke \bgroup \em et al.\egroup
  }{1997}]{wilke-etal97}
W.~Wilke, I.~Vollrath, K.-D. Althoff, and R.~Bergmann.
\newblock A framework for learning adaptation knowledge based on knowledge
  light approaches.
\newblock In {\em Proceedings of the Fifth German Workshop on Case-Based
  Reasoning}, pages 235--242, 1997.

\bibitem[\protect\citeauthoryear{Wilson and
  O'Sullivan}{2008}]{wilson-osullivan08}
D.~Wilson and D.~O'Sullivan.
\newblock Medical imagery in case-based reasoning.
\newblock In P.~Perner, editor, {\em Case-Based Reasoning on Images and
  Signals}, volume~73 of {\em Studies in Computational Intelligence}, pages
  389--418. Springer, 2008.

\bibitem[\protect\citeauthoryear{Wiratunga \bgroup \em et al.\egroup
  }{2006}]{craw-wiratunga-rowe06}
N.~Wiratunga, S.~Craw, and R.~Rowe.
\newblock Learning adaptation knowledge to improve case-based reasoning.
\newblock {\em Artificial Intelligence}, 170:1175--1192, 2006.

\bibitem[\protect\citeauthoryear{Ye \bgroup \em et al.\egroup
  }{2021}]{ye-et-al21}
X.~Ye, D.~Leake, V.~Jalali, and D.~Crandall.
\newblock Learning adaptations for case-based classification: A neural network
  approach.
\newblock In {\em Case-Based Reasoning Research and Development, ICCBR 2021}.
  Springer, 2021.
\newblock In press.

\end{thebibliography}
\end{document}